\pdfoutput=1

\documentclass[11pt]{article}

\usepackage[]{emnlp2021}

\usepackage{times}
\usepackage{latexsym}

\usepackage[T1]{fontenc}

\usepackage[utf8]{inputenc}

\usepackage{microtype}
\usepackage{times}
\usepackage{epsfig}
\usepackage{graphicx}
\usepackage{amsmath}
\usepackage{amssymb}

\usepackage{subcaption}
\usepackage{booktabs}
\usepackage{multirow}
\usepackage{enumitem}
\usepackage{algorithmic}
\usepackage{algorithm}
\usepackage{xcolor}

\newtheorem{theorem}{Theorem}

\newtheorem{definition}[theorem]{Definition}
\newcommand{\gpt}{\text{GPT-3}}
\newcommand{\err}{\text{err}}
\title{ 
Want To Reduce Labeling Cost? GPT-3 Can Help 
}

\author{
Shuohang Wang \quad Yang Liu \quad Yichong Xu \quad Chenguang Zhu \quad Michael Zeng\\ \\
    Microsoft Cognitive Services Research Group \\
    {\small \{\tt shuowa,yaliu10,yicxu,chezhu,nzeng\}@microsoft.com} 
}

\begin{document}
\maketitle
\begin{abstract}

Data annotation is a time-consuming and labor-intensive process for many NLP tasks. Although there exist various methods to produce pseudo data labels, they are often task-specific and require a decent amount of labeled data to start with. Recently, the immense language model GPT-3 with 175 billion parameters has achieved tremendous improvement across many few-shot learning tasks.
In this paper, we explore ways to leverage GPT-3 as a low-cost data labeler to train other models. We find that, to make the downstream model achieve the same performance on a variety of NLU and NLG tasks, it costs 50\% to 96\% less to use labels from GPT-3 than using labels from humans.
Furthermore, we propose a novel framework of combining pseudo labels from GPT-3 with human labels, which leads to even better performance with limited labeling budget.
These results present a cost-effective data labeling methodology that is generalizable to many practical applications.

\end{abstract}

\section{Introduction}
Data always plays a crucial role in developing machine learning models. However,  collecting human-labeled data is a costly and time-consuming process, especially in multi-task scenarios. 
With the success of pre-trained models~\cite{zhang2020pegasus,raffel2019exploring,liu2019roberta,devlin2018bert} on unlabeled data, the performance of models under few-shot and zero-shot settings has been greatly enhanced.
In particular, the large-scale language model GPT-3~\cite{brown2020language}, with 175 billion parameters, is the state-of-the-art few shot learner on many NLP tasks.

However, GPT-3 is constrained on its immense model size and requires a large amount of resource to be deployed for real applications. 
Moreover, GPT-3 doesn't provide a free lunch, and its public API has a charge correlated with the number of processed tokens\footnote{\url{https://beta.openai.com/pricing}}. Thus, an interesting problem arises: instead of directly deploying GPT-3 for downstream tasks, how can we leverage GPT-3 to achieve a more cost-effective and efficient training of other models?

In this paper, we employ GPT-3 to label unannotated data to train smaller models which are deployed for inference. Although the data labeled by GPT-3 is usually more noisy than human-labeled data, the process is much cheaper, faster and generalizable to multiple tasks. For example, for the Stanford Sentiment Treebank (SST-2) task~\cite{sst}, it takes as low as 0.002 dollars on average to use the GPT-3 API to annotate one label. However, it costs 0.11 dollars to label an instance on crowd-sourcing platforms. Plus, the GPT-3 API can label data non-stoppingly at a much faster speed than human labelers.

In our extensive empirical analysis, we find that to make in-house models (e.g. PEGASUS~\cite{zhang2020pegasus}, RoBERTa~\cite{liu2019roberta}) to achieve the same performance on various NLU and NLG tasks, data labeled by GPT-3 incurs a much lower cost (e.g. 50\%-95\% lower) than data labeled by humans, especially in low-resource settings. Moreover, we also find that these in-house models trained with data labeled by GPT-3 can outperform GPT-3 itself under the fewshot setting, which we give theoretical justifications.

\begin{figure*}[!ht]
  \centering
  \includegraphics[width=\linewidth]{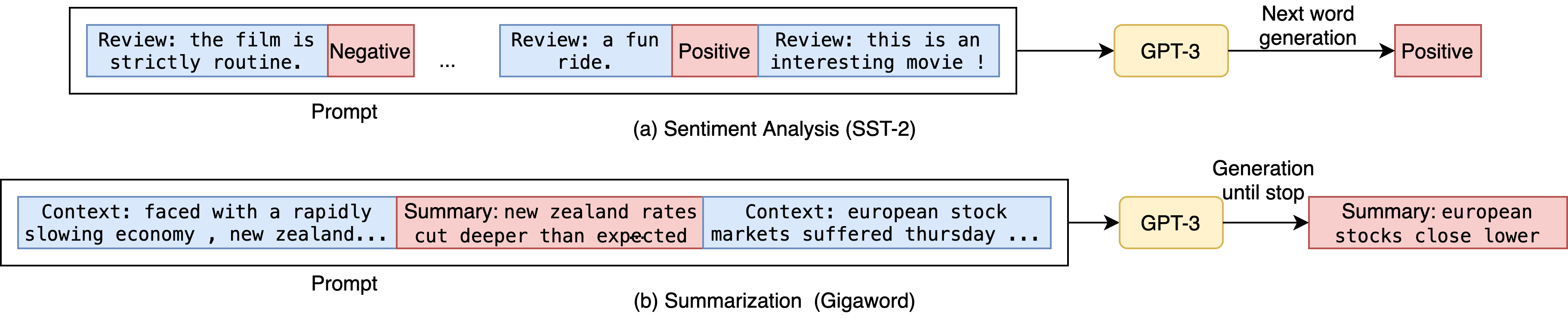}
  \caption{Two examples of constructing GPT-3 input. The input prompt of GPT-3 consists of $n$ labeled data ($n$-shot learning) and the task input for which GPT-3 generates the label. The same $n$ labeled data is used for every input. 
  }
  \label{fig:prompt}
\end{figure*}
In addition to using labeled data from a single source, we explore ways to smartly assign unlabeled data to different labelers, i.e. GPT-3 and human, under a fixed budget. We frame this as a dual supervision problem~\cite{jung2020dual} with cost and budget constraints. In detail, we tried mixing data labeled by GPT-3 and humans with different ratios: 25\%, 50\%, 75\% of the budget. Moreover, we propose an active labeling strategy to have humans re-annotate data labeled by GPT-3 with the lowest confidence scores. Both strategies manifest clear improvement over using a single source of labeler.

We conduct comprehensive empirical analysis of our proposed cost-effective labeling strategies on 9 NLP tasks, including text entailment~\cite{rte,de2019commitmentbank}, sentiment analysis~\cite{sst}, topic classification~\cite{agnewsdbpedia}, answer type classification~\cite{trec}, summarization~\cite{gigaword,xsum}, and question generation~\cite{rajpurkar2016squad}. We show that our labeling strategy can significantly reduce labeling cost while achieving the same performance with human-labeled data. For instance, our method saves 96\% cost on the sentence classification task SST-2, 93.8\% cost on the summarization task Gigaword, and 50-75\% cost on other tasks.

We summarize our contributions as follows:
\begin{enumerate}
\setlength{\itemsep}{0.5pt}
    \item We propose to leverage GPT-3 as a data labeler which can save 50\% to 96\% cost to achieve the same performance compared with human labeling, on a variety of NLP tasks.
    \item We observe that the in-house models (e.g. PEGASUS, RoBERTa) trained on GPT-3 labeled data can outperform the GPT-3 fewshot learner. 
    \item We explore various strategies of mixing labeled data from GPT-3 and humans under a fixed budget and achieve better performance than using data from a single labeler.
    \item We propose a novel active labeling method to have human labeler re-annotate data from GPT-3 with lowest confidence score. 
    \item To the best of our knowledge, this is the first work to analyze the cost of GPT-3 in data labeling and the effect of mixing data labeled from GPT-3 and humans.
\end{enumerate}

\section{Method}
\begin{figure*}[!th]
  \centering
  \includegraphics[width=\linewidth]{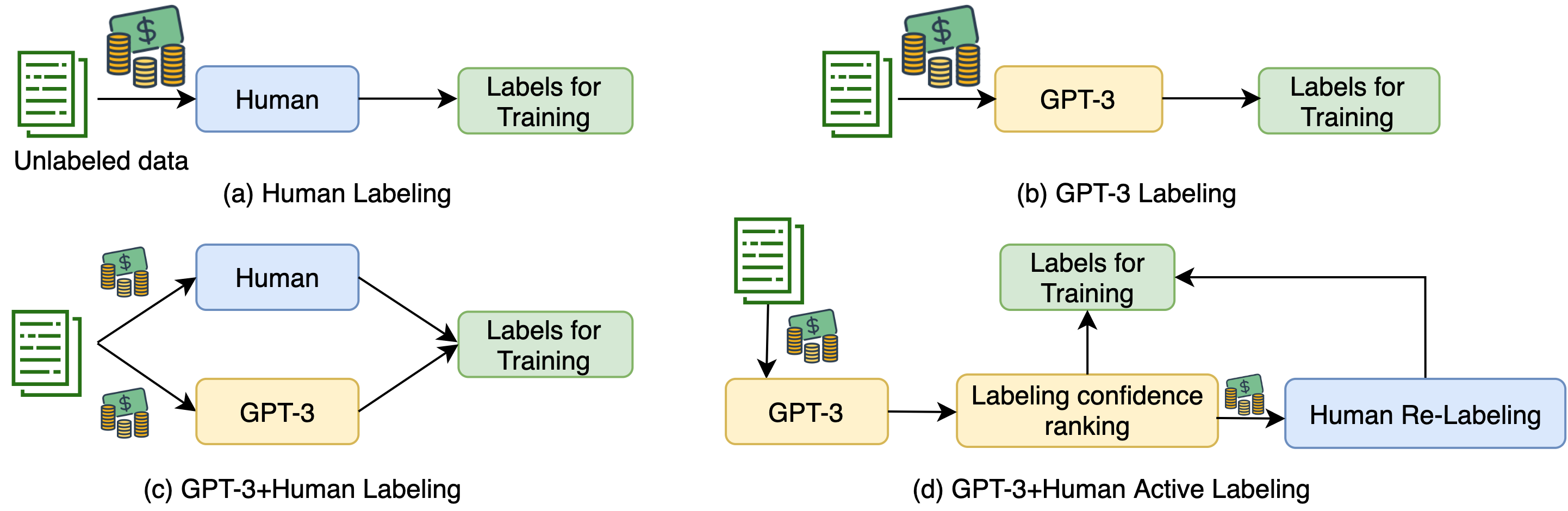}
  \caption{Four data labeling strategies given a fixed budget. a) label data by human only, b) label data by GPT-3 only, c) randomly select non-overlapped data according to a split ratio of budget for human and GPT-3 to label, d) select GPT-3 labeled data with lower confidence scores for humans to re-label. }
  \label{fig:mixup}
\end{figure*}
In this section, we introduce how GPT-3 can help reduce labeling costs. First, we present a cost analysis of GPT-3 and human labeling. Next, we introduce how to use GPT-3 to label unannotated data. Then, we theoretically explain why a downstream model trained with GPT-3 labels can outperform GPT-3 itself. Finally, we show how to mix up labels from GPT-3 and humans to further boost performance at a lower cost.

\subsection{Labeling Cost Analysis}
\begin{table}[]
\small
\begin{tabular}{lccccc}
\toprule
         & \multicolumn{1}{l}{\#Tok} & \multicolumn{3}{c}{GPT-3} & Human \\
NLG         &                            & 1-Shot   & 2-Shot  & 3-Shot  &       \\ \midrule
Gigaword & 31                         & 2.5e-3  & 3.7e-3 & 5.0e-3 & 0.11  \\
SQuAD    & 126                        & 1.0e-2  & 1.5e-2 & 2.0e-2 & 0.28  \\
XSum     & 382                        & 3.5e-2  & 4.6e-2 & 6.1e-2 & 0.84  \\ \midrule
NLU         &                            & 2-Shot   & 4-Shot  & 8-Shot  &       \\ \midrule
SST-2    & 19.3                       & 2.3e-3  & 3.9e-3 & 6.9e-3 & 0.11  \\
CB       & 62.7                       & 7.5e-3  & 1.2e-2 & 2.3e-2 & 0.11  \\
TREC     & 10.2                       & 1.2e-3  & 2.0e-3 & 3.6e-3 & 0.11  \\
AGNews   & 31.6                       & 3.8e-3  & 6.3e-3 & 1.1e-2 & 0.11  \\
DBPedia  & 47.3                       & 5.7e-3  & 9.5e-3 & 1.7e-2 & 0.11  \\
RTE      & 52.4                       & 6.3e-3  & 1.2e-2 & 1.9e-2 & 0.11 \\ \bottomrule
\end{tabular}
\caption{Cost (\$) per GPT-3 and Human labeling. \#Tok is the number of tokens on average from the corresponding dataset. For different GPT-3 few-shot labeling strategies, it charges differently based on the sequence length. The final cost per label for n-shot GPT-3 is $\#tok\times 4\times 10^{-5}\times(n+1)$, where $4\times 10^{-5}$ is the cost GPT-3 charged per token. For human labeling, it costs \$0.11 per 50 input tokens with a minimum of \$0.11.}
\label{tbl:cost}
\end{table}
In this section, we compare the costs of GPT-3 and crowd-sourced labeling. To make it simplified, we ignore the cost for GPT-3 template selection, human labeler selection, etc., and only consider the labeling cost charged per label from API or crowd-sourcing platform. We show a detailed comparison in Table~\ref{tbl:cost}. 

\paragraph{Cost of GPT-3 labeling.} 
The GPT-3 API provided by Open-AI charges by the number of tokens to encode and generate. We get the quotes from Open-AI, ``2M tokens for \$100 per month, 10M tokens for \$400 per month, or Contact Us for larger scale”. We use the \$400 quote for all our experiments. As the sequence length of different datasets can be significantly different, it costs differently to label one instance by GPT-3 (Table~\ref{tbl:cost}). Moreover, different GPT-3 few-shot labeling strategies are also charged differently. More shots lead to a higher cost per GPT-3 labeling as the prompt is longer. 

\paragraph{Cost of human labeling.} We estimate the crowd-sourcing labeling price from Google Cloud Platform\footnote{\url{https://cloud.google.com/ai-platform/data-labeling/pricing\#labeling\_costs}}. For labeling classification tasks, it charges 1000 units (50 tokens per unit) for \$129 in Tier 1 and \$90 in Tier 2. We adopt the average cost from Tier 1\&2 as the human labeling cost. For generation tasks, there is no detailed instruction, as the rate can be quite different based on task difficulty. Thus, we follow the cost of classification tasks by charging \$0.11 per 50 tokens. Here, we note that the actual human labeling is often more expensive. For example, the same instance is labeled by multiple labelers for majority voting; some datasets are labeled by experts, not by crowd-sourcing.

Overall, GPT-3 can be more than ten times cheaper than human labeling on average, making GPT-3 label much more data than human under the same budget. Moreover, we believe in the future GPT-3 API price will likely drop as better technologies emerge, while human labeling price is likely to stay the same or become even more expensive. 

\subsection{GPT-3 Labeling}
GPT-3~\cite{brown2020language} is a large-scale pre-trained language model, and we use the largest  model, Davinci, from OpenAI to label data. 
Given a sequence, GPT-3 can generate output that naturally follows the input. According to the GPT-3 API from OpenAI, we can feed it an input sequence with up to 2,048 tokens. The output is a sequence ending with a special stop sign. At the meantime, the API returns the logits for top-k predicted tokens at each output position. 

We propose to use this GPT-3 API for data labeling. An overview of the process is shown in Figure~\ref{fig:prompt}.

Here, we formulate the GPT-3 labeling process as follows:
\begin{equation}
    \text{Y}_i, \text{logit}_i = \text{GPT-3} (\text{Labeled-Data}, X_i)
    \label{eqn:gpt}
\end{equation}
where $\text{Y}_i$ is a textual sequence with $l$ tokens,  $\text{logit}_i\in \mathbb{R}^l$ is the corresponding logits. The input sequence to GPT-3 consists of two parts: several human-labeled textual sequences and a target input sequence at the end, $X_i$. 

The label collection from the GPT-3 output depends on the task type. For classification tasks, we only collect the first output token which is the label, e.g. Positive or Negative\footnote{We use the bias option in GPT-3 API to limit the output token to be within the set of label text.}.
For generation tasks, we collect the entire output as the label. 

As the cost from GPT-3 API is computed based on length of input sequence plus that of the output, we consider variants of input sequences. $n$-shot GPT-3 means we place $n$ human-labeled instances in the input prompt, of which the cost is included. When $n$ is smaller, the overhead of human labels is cheaper, as well as the labeling cost of GPT-3. For instance, in SST-2, using 8-shot GPT-3 to label is about 4.5 times more expensive than using 1-shot GPT-3. However, a larger $n$ would usually lead to better labeling quality. So it is a trade-off according to the labeling budget. In this paper, we explore 2,4,8-shots for NLU tasks and 1,2,3-shots for NLG tasks.

After we collect labels for unannotated data from GPT-3, we train smaller in-house model on the tasks: PEGASUS \citep{zhang2020pegasus} for NLG tasks and RoBERTa$_{large}$ \citep{liu2019roberta} for NLU tasks.

\subsection{Is Using GPT-3 Labeling Better Than GPT-3 Itself?}
\label{sec:better}
\citet{brown2020language} propose to directly use GPT-3 for downstream tasks, with the $n$ given labeled instances and no fine-tuning. We refer to this strategy as raw GPT-3. 

We note that raw GPT-3 is expensive, as its cost goes linearly with the number of instances during inference. Also, it has a relatively high latency when deployed for real applications. 

However, even in terms of accuracy, we observe in the experiments from section~\ref{sec:result} 
that the in-house models trained with GPT-3 labels can often outperform raw GPT-3. 
We argue that by using data labeled by GPT-3, we are essentially performing \emph{self-training}: the predictions on unlabeled samples act as regularization on induced models and help improve the performance. In particular, for classification problems, we can theoretically upper-bound the error rate of the best in-house model using the labels generated by GPT-3. 

\begin{definition}[Consistency assumption]
Define $\mathcal{X}$ as the input space and $\mathcal{G}$ as the set of classifiers we train. The consistency assumption says that $\exists r>0$, such that $\forall G\in \mathcal{G}, \forall x,x' \in \mathcal{X}$, if $x'\in B(x)=\{x': \|x'-x\|\leq r \}$, we have $G(x')=G(x)$.
\end{definition}

Under this consistency assumption, we can follow previous theoretical results~\citep{wei2020theoretical} to show the following:

\begin{theorem}
\label{thm:self-training}
Suppose $\hat{G} \in \mathcal{G}$ is the classifier that minimizes its discrepancy with GPT-3 over the input space $\mathcal{X}$. Let $\bar{a}$ be the maximum error of GPT-3 on any class $P_i$. If $P$ satisfies $(\bar{a}, \bar{c})$-expansion, then we have
\[\err(\hat{G}) \leq \frac{2}{c-1}\err(\gpt),\]
where $c=\min\{1/\bar{a}, \bar{c}\}$.
\end{theorem}
Here $c>3$ is a distribution-dependent constant. We provide the definition of expansion along with the proof in the appendix.
Thus, it shows that the error rate of our trained $\hat{G}$ using GPT-3 labels can be lower than that of GPT-3 itself.

\begin{figure*}[!t]
  \centering
  \includegraphics[width=\linewidth]{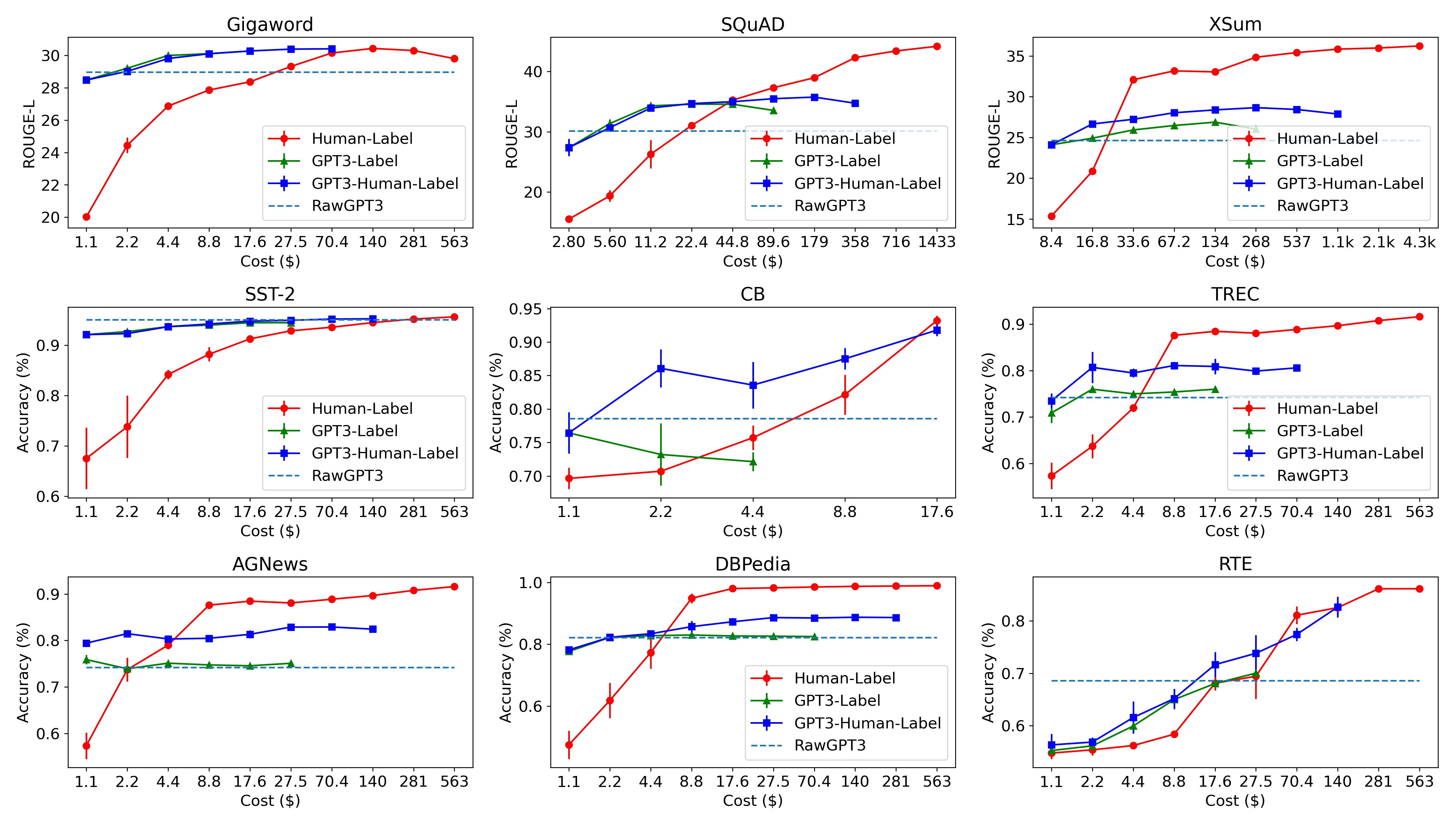}
  \caption{Performance v.s. labeling cost of various labeling strategies on 9 NLG and NLU datasets. X-axis is the cost in dollar estimated by OpenAI pricing policy and crowd-sourced annotation.
  Each point is the average result of 3 runs of PEGASUS (NLG) or RoBERTa$_{large}$ (NLU) using 3 sets of generated labels, with the standard deviation shown. The performance of using GPT-3 as the inference model is shown as a dashed line, which is the maximum ROUGE-L/accuracy over different shot settings. Note that the cost of GPT3-Label and GPT3-Human-Label cannot further increase when all training data (up to 5,120 instances) has been labeled.}
  \label{fig:gpthuman}
\end{figure*}
\subsection{GPT3-Human Labeling}
Although labels from humans are more expensive, they are often of a higher quality than GPT-3 labels. Thus, we explore ways to mix labels from both human and GPT-3 to reduce cost and improve performance. 

Given a fixed budget, we split it for labeling by humans and GPT-3, as shown in Figure~\ref{fig:mixup} (c). 
In this way, the in-house model is exposed to data from both sources. So the training loss is in the form of dual supervision on two disjoint sets of labeled data as follows:
\begin{equation}
    L=\sum_{i\in T}L_g(Y_i, X_i) + \alpha \sum_{j\in H}L_h(Y_j, X_j)
    \label{eqn:loss}
\end{equation}
where $T$ is a set of GPT-3 labeled data,  $H$ is a set of human labeled data, and their sizes depend on the budget split ratio. In out experiments, we try to assign 0\%, 25\%, 50\%, 75\%, and 100\% of budget to each type of labeling. Considering GPT-3 labels may be noisier than human labels, we also add a weight $\alpha$ between two types of supervision. As the unlabeled data are randomly assigned to GPT-3 or human, we refer to this GPT3-Human strategy as \textit{random labeling}.

\paragraph{Active labeling}
GPT-3 API provides logits together with the generated text (Equation~\ref{eqn:gpt}). For NLU tasks, we treat the logit of the first generated word as the confidence score for this label. In experiments, we observe a high correlation between the accuracy of GPT-3 labels and these confidence scores (Figure~\ref{fig:active}).

Thus, a question naturally arises: can we leverage the high quality of human labeling to help re-annotate these low-quality labels?

We therefore propose an active labeling method for NLU tasks to have humans re-annotate GPT-3 labels for which the uncertainty is the highest (Figure~\ref{fig:mixup} (d)). In detail, GPT-3 first labels the data. Then, we rank all the labels based on the confidence score (logit) and select those with the lowest scores to be re-labeled by humans. All the budget for human labeling is dedicated to this relabeling. In our experiments, the number of data to label depends on the budget assigned to either GPT-3 or human, and we will show different strategies to split the budget. Finally, the relabeled data and other GPT-3 labeled data are fed into in-house models for fine-tuning.

\section{Experiments}

\begin{figure*}[!ht]
  \centering
  \includegraphics[width=\linewidth]{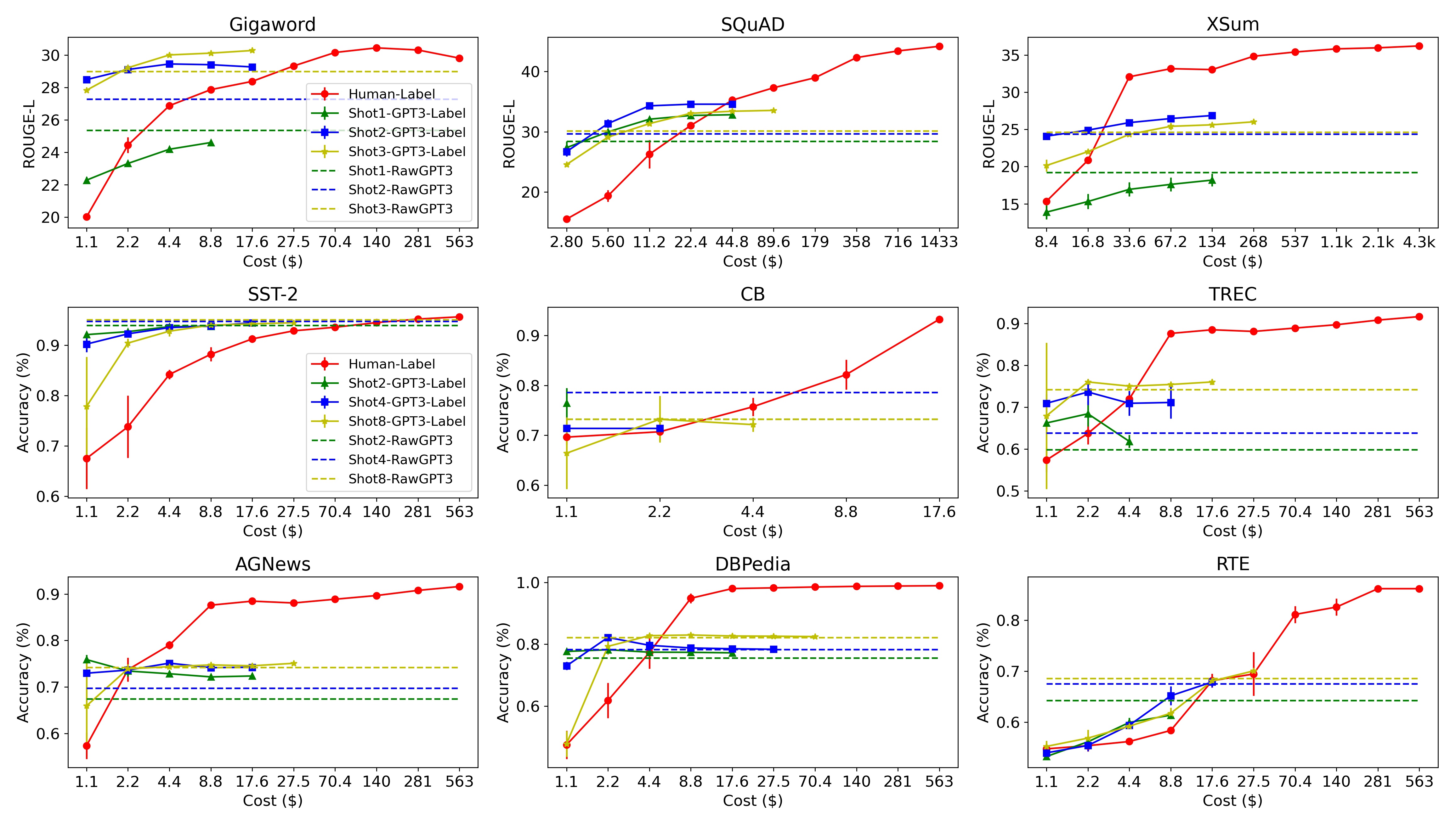}
  \caption{GPT-3 labeling performance. We feed un-labeled data to GPT-3 with different shot settings and fine-tune Transformer models on the corresponding labeled data. The dot lines are the raw GPT-3 performance with various shots. Lines in the same color use the same number of shots in GPT-3. The cost of GPT3-Label cannot further increase when all training data (up to 5,120 instances) has been labeled.}
  \label{fig:gptonly}
\end{figure*}
\subsection{Datasets}
We employ 3 natural language generation (NLG) tasks and 6 natural language understanding (NLU) tasks for evaluation. We sample up to 5.1K cases from the training data for labeling. 
We simulate human labeling by using the labels from the datasets.
We use the original test set for evaluation if it is available, and use development set otherwise.

\paragraph{NLG tasks} 
We apply our labeling strategies to natural language generation tasks, two on summarization and one on question generation task.
\textbf{XSum}~\cite{xsum} is from BBC articles, each of which contains an expert-written summary. 
\textbf{Gigaword}~\cite{gigaword} also comes from news articles, and the task is to summarize the first sentence in the article by generating its headline.
\textbf{SQuAD}~\cite{rajpurkar2016squad} is Stanford Question Answering dataset, and our task is to generate a question given a paragraph and an answer.

\paragraph{NLU tasks}
We leverage the following classification tasks. \textbf{SST-2}~\cite{sst} is a binary sentiment classification task from Stanford Sentiment Treebank.
\textbf{TREC}~\cite{sst} is to identify an answer type of a question from Number, Location, Person, Description,
Entity, or Abbreviation.
\textbf{CB}~\cite{de2019commitmentbank} is a 3-way textual entailment task to classify a sentence pair of premise and hypothesis into Contradiction, Entailment, or Neutral. 
\textbf{RTE}~\cite{rte} is a 2-way text entailment: Entailment or Not-Entailment.  
\textbf{AGNews}~\cite{agnewsdbpedia} is to identify the topic from World, Sports, Business, and Technology. 
\textbf{DBPedia}~\cite{agnewsdbpedia} provides a different topic pool: Company, School, Artist, Athlete, Politician, Transportation, Building, Nature,
Village, Animal, Plant, Album, Film, or
Book. 
\subsection{Settings}
\label{sec:setting}

\paragraph{Model structure} For GPT-3 labeling API, we select the largest version Davinci\footnote{\url{https://beta.openai.com/pricing}}. Our in-house NLG model is initialized by PEGASUS$_\text{large}$~\cite{zhang2020pegasus} which is a Transformer with 16 encoder and decoder layers, 1024 hidden size, and 16 attention heads. 
Our in-house NLU model is initialized by RoBERTa$_\text{large}$~\cite{liu2019roberta} which is a Transformer with 24 encoder layers, 1024 hidden size, and 16 attention heads. Our fine-tuning codes are mainly  based on Hugging Face Transformer library\footnote{\url{https://github.com/huggingface/transformers}}.

\paragraph{Labeling strategy} We evaluate 3 categories of labeling strategies: 1) fully human labeling, 2) fully GPT-3 labeling, 3) GPT-3 and human mix-up labeling. Within each category, the hyper-parameters include: 1) number of GPT-3 shots, \{1,2,3\} shots for NLG tasks and \{2,3,4\} for NLU tasks, 2) GPT-3 and human labeling mix-up budget ratio chosen from \{0\%, 25\%, 50\%, 75\%, 100\%\}, 3) labeling method when mixing GPT-3 and human labeling, \{random labeling, active labeling\}, where random labeling means there is no human re-labeling. For each strategy, we try 3 seeds to shuffle the data to label. The budget limits are set to the cost of human labeling 10, 20, 40, 80, 160, 320, 640, 1,280, 2,560 and 5,120 samples in each dataset (Table~\ref{tbl:cost}).
 
 \paragraph{Fine-tuning} For fine-tuning both NLG and NLU tasks, the hyper-parameters are searched from learning rate \{1e-5, 3e-5\}, 
 batch size \{8, 32\}, epochs \{3,7,20\},
 weight $\alpha$ \{1,3\} in Eqn.(\ref{eqn:loss}) on human labels.  
 
 \subsection{Experiment Result}
 \label{sec:result}
 
\begin{figure*}[!ht]
  \centering
  \includegraphics[width=\linewidth]{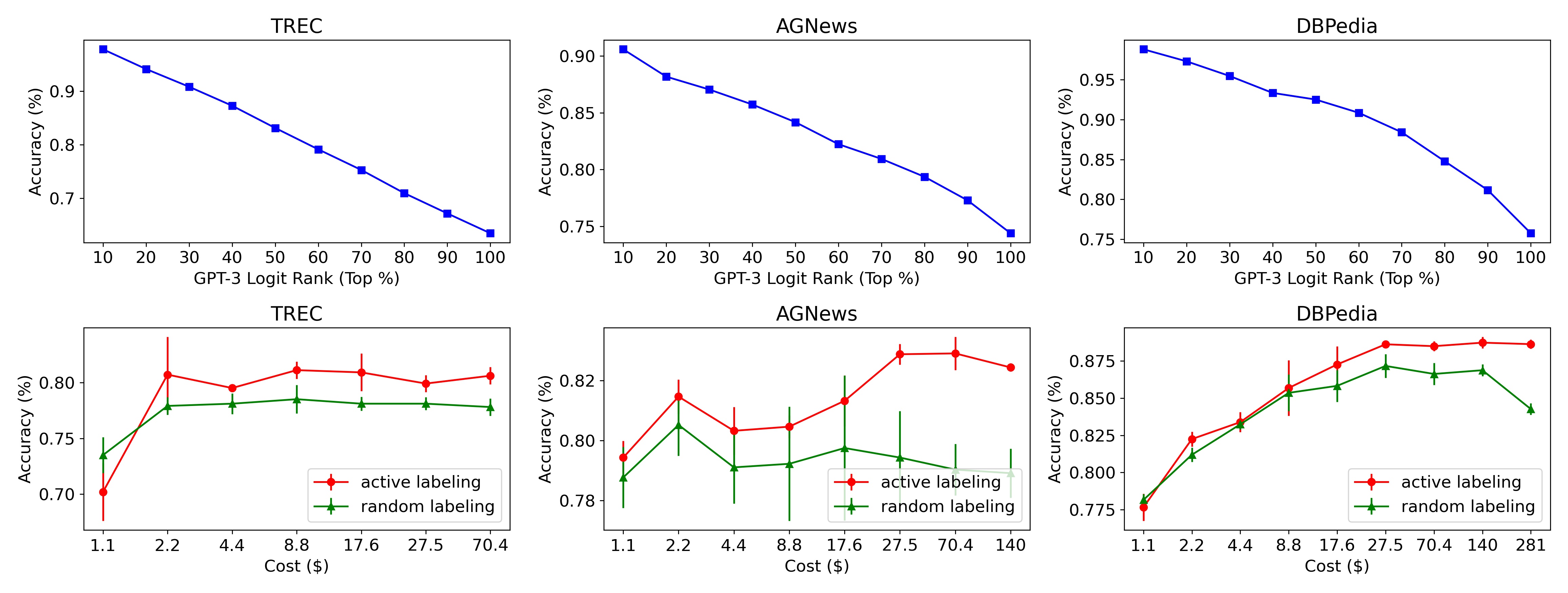}
  \caption{Active labeling. The first row shows that logit values from GPT-3 can be treated as confidence scores, and high-confidence labels are much more accurate than low-confidence ones. 
  The second row compares the performance of active labeling and random labeling in GPT3-Human strategy on three different NLU datasets. }
  \label{fig:active}
\end{figure*}
 \subsubsection{Main Result} 

 In Figure~\ref{fig:gpthuman}, we are trying to identify which labeling strategy has potential to work best with a fixed budget: fully human labeling, fully GPT-3 labeling, or GPT3-Human mix-up labeling? 
 The experiment results are the max value over different labeling hyper-parameters, as described in Section \ref{sec:setting}, and we report the mean and standard deviation of 3 trials.
 From the figure, we can see that for all tasks, fully GPT-3 labeling can achieve better performance than fully human labeling in low-budget settings, and GPT3-human mix-up labeling can further improve the performance.

 For most tasks except RTE, with only \$1.1 budget, GPT-3 based labeling can already lead to a good dataset for fine-tuning. For instance, in SST-2, RoBERTa trained with GPT-3 labeled data under a budget of \$1.1 can achieve the same performance with using human labels worth \$27.5, with a 96\% saving of labeling cost. For the summarization task Gigaword, PEGASUS trained with GPT-3 labels of \$4.4 budget can achieve the same performance with using human data worth \$70.4, a saving of 93.8\%. 
 
 Overall, we observe a 50\%-96\% of cost saved by GPT-3 labeling (fully GPT-3 and GPT3-Human mix up) to achieve the same performance as using human labels, under low-budget settings. We note that with the fast development of infrastructure and more advanced algorithms, the cost of GPT-3 API will likely reduce in the future, making our labeling strategies even more attractive.
 
 Also, we observe that when the budget is ample or unlimited, fully human labeling will dominate in performance due to higher quality. However, when the budget is limited, GPT-3 labeling is a more cost-effective choice.
 
\subsubsection{GPT-3 Labeling}

Figure~\ref{fig:gptonly} shows the performance of GPT-3 labeling under different few-shot settings and that of raw GPT-3. 
For most NLU datasets, e.g. SST-2, TREC, AGNews, and DBPedia, fewer shot GPT-3 labeling can lead to better performance. The main reason is that 2-shot GPT-3 labeling is much cheaper than 8-shot and can label much more data under the same budget. But when the budget further increases, the labeling quality comes to be a pivotal factor for better performance. For NLG datasets of Gigaword and XSum, the performance of 1-shot GPT-3 labeling is much worse than that of 2-shot and 3-shot, due to lower label qualities.

We also observe that the in-house models trained with enough GPT-3 labels outperform raw GPT-3 (dotted lines with the same color). It shows that our GPT-3 labeling strategy can not only be treated as a cost efficient self-annotation method, but also a semi-supervised method to further boost performance of few-shot learners.

\subsubsection{Active Labeling} 
Recall that active labeling is used in GPT3-Human strategy, in which humans re-label the low-confidence instances given by GPT-3.
The first row of Figure~\ref{fig:active} shows there is a strong correlation between the accuracy of GPT-3 labels and its confidence score, represented by the logit returned by the API. For instance, the GPT-3 labels with top 10\% logits have an accuracy of 95\%, 90\%, 95\% for TREC, AGNews, and DBPedia respectively, while low-confidence labels have a much lower accuracy. 
As a result, active labeling can help improve the quality of labels, which leads to better performance of downstream models, as shown in the second row of Figure~\ref{fig:active}. For example, in TREC, active labeling can boost the accuracy from 77\% to 80\% under the same budget of \$2.2. With active labeling, we also work on a real strategy of mixing GPT-3 and human labeling by equally splitting the budget. We also have done experiments with different shots for GPT-3. The final curve of performance v.s. labeling cost of this strategy is quite similar to Figure~\ref{fig:gptonly}. Thus we leave it in Appendix~\ref{app:real} for reference.

\section{Related Work}
\paragraph{GPT-3 Overview.} With the success of large pre-trained language modeling GPT-3~\cite{brown2020language} on few-shot learning,  more works have been done to improve GPT-3. \citet{zhao2021calibrate} propose to remove the model bias before using GPT-3, which not only increases the accuracy but also reduces the variance.
 \citet{lu2021fantastically} work on how to order the few labeled data as input of GPT-3 by constructing
an artificial development set. One concurrent with our work, \citet{yoo2021gpt3mix} consider distilling knowledge from GPT-3 with synthetic data. In their work, the synthetic dataset size is always the same as the original training dataset size.  Unlike the most recent works on GPT-3, we treat GPT-3 as a new source of labeler and focus on analyzing the cost of running GPT-3, which is not free according to OpenAI API.
This work is complementary to many other methods based on human labeling, such as few-shot learning~\cite{yin2020meta}, active learning \citep{settles2009active, dor2020active} and transfer learning \cite{ruder2019transfer}. 

\paragraph{Dual supervision.} Our method is also related to dual supervision \cite{attenberg2010unified}, which combines two types of labels (one cheap and one expensive) to train a model. Dual supervision typically considers different labeling tasks for humans, for example labeling words or documents \citep{melville2009active}, natural language understanding or generation \citep{su2019dual}, cardinal or ordinal labels \cite{xu2020regression}; here, we consider the same task for different-cost labelers.  Labeling oracles with different costs for the same task have also been considered in other areas. Proactive learning \citep{donmez2008proactive} considers active learning with multiple oracles with varied label quality and cost, and oracles can also abstain from labeling an example (``unknown'' label). Multi-fidelity optimization \citep{song2019general} considers optimizing an underlying function (e.g., development accuracy of a neural network) by querying approximations of different precisions and costs. 

\paragraph{Semi-supervised learning and Self Training.}

Using existing model predictions for semi-supervised learning is well-explored in self-training \cite{yarowsky1995unsupervised,mukherjee2020uncertainty}. Prior works in self-training has achieved state-of-art performance in tasks like machine translation \cite{he2019revisiting} and task-oriented dialogue understanding \cite{wang2020adaptive}.
However, prior works in self-training typically used similar-sized models for teacher and student, where the cost of obtaining labels from the teacher is negligible. Learning from GPT-3 is particularly promising because of its impressive few-shot performance, but also challenging because of the GPT-3 labeling cost. To the best of our knowledge, this is the first work that explicitly considers the cost of GPT-3 and its effect in reducing the labeling cost.

\section{Conclusion} 
In conclusion, we investigate how to use GPT-3 to label unannotated data in a cost-efficient way. We show that our strategies can significantly reduce the labeling cost by achieving the same performance with human-labeled data. We also find that models trained with GPT-3 labels can achieve better performance than raw GPT-3. Moreover, we introduce the GPT3-Human labeling strategy, which outperforms both fully human and fully GPT-3 labeling. Finally, we propose active labeling to leverage the advantages from human and GPT-3, which works better than randomly selecting data to label on multiple NLP tasks. Our work shows the potential in cost-efficient data labeling with few-shot learners.

For future work, we plan to extend our methods to data augmentation to produce both instances and labels.

And it is worth noting that GPT-3 is not reliable enough yet at labeling ``high-stakes" cases, e.g. identifying toxic language, but is more suitable for low-stakes labeling\footnote{\url{https://beta.openai.com/docs/safety-best-practices}}.

\bibliography{anthology,custom}
\bibliographystyle{acl_natbib}

\appendix
\newpage
\onecolumn
\section{Proof of Theorem \ref{thm:self-training}}
\label{sec:proof_theorem}
We follow \citet{wei2020theoretical} and use their definition of expansion:
\begin{definition}[$(a, c)$-expansion, \citet{wei2020theoretical}]
Let $P$ be the sample distribution, and $P_i$ be the class-conditional distribution $P(X|\text{label}(X)=i)$. We say that the class-conditional distribution Pi satisfies $(a, c)$-expansion if for all set $V$ with class probability $P_i(V)\leq a$, the following holds:
\[P_i(N(V)) \geq \min\{cP_i(V), 1\},\]
where $N(V)$ is a distribution-dependent neighborhood of $V$ (see \citet{wei2020theoretical} for details). If $P_i$ satisfies $(a, c)$-expansion for all label $i$, then we say $P$ satisfies $(a, c)$-expansion.
\end{definition}
Please refer to \citet{wei2020theoretical} for theoretical and experimental justification of the expansion property. 

\noindent\textbf{Proof of Theorem \ref{thm:self-training}} Our theorem is a direct consequence of Theorem 4.3 in \citet{wei2020theoretical}. Our consistency assumption leads to the condition of $R_B(G)=\mu=0$ for any classifier $G$ we consider, in Theorem 4.3 and (4.1) of \citet{wei2020theoretical}. This directly proves our Theorem \ref{thm:self-training}.


\section{GPT-Human Labeling}
\label{app:real}
\begin{figure*}[!ht]
  \centering
  \includegraphics[width=\linewidth]{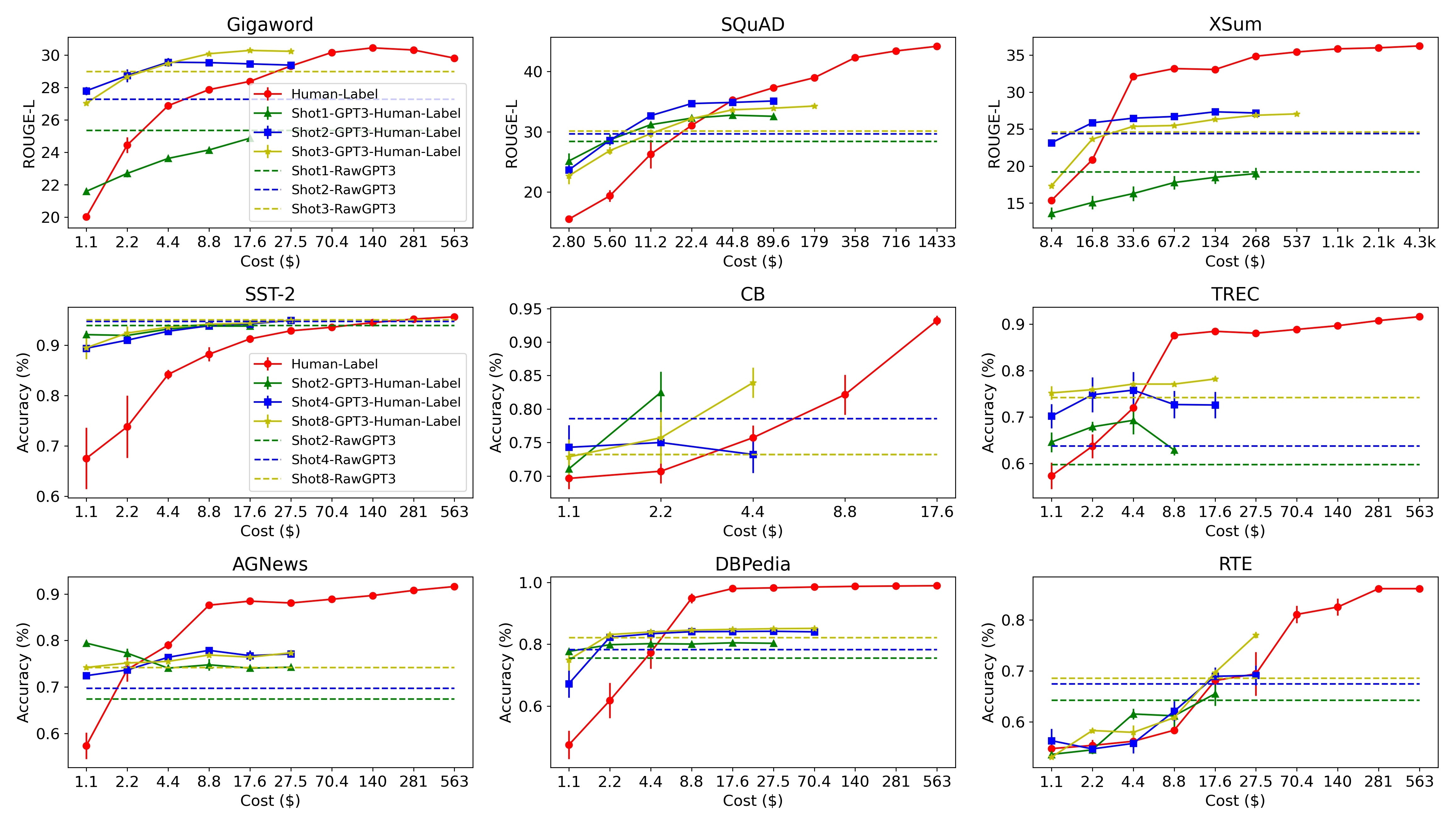}
  \caption{GPT3-Human labeling performance. The budget is equally split for GPT3 and human labeling. Active labeling is adopted in this experiment.}
  \label{fig:realstrategy}
\end{figure*}

\end{document}